\documentclass{article}
\usepackage{spconf,amsmath,epsfig}

\usepackage{booktabs}
\title{SAR IMAGE DESPECKLING THROUGH CONVOLUTIONAL NEURAL NETWORKS}
\twoauthors
  {G. Chierchia}
  {Universit\'e Paris Est\\
  	LIGM UMR 8049, CNRS, ESIEE Paris\\
  	F-93162 Noisy-le-Grand (France)}
  {D. Cozzolino, G. Poggi, L. Verdoliva}
	{University Federico II of Naples\\
	DIETI\\
	Via Claudio 21, 80125 Naples (Italy)}
\begin{document}
%
\maketitle
\begin{abstract}
In this paper we investigate the use of discriminative model learning through Convolutional Neural Networks (CNNs) for SAR image despeckling.
The network uses a residual learning strategy,
hence it does not recover the filtered image,
but the speckle component, which is then subtracted from the noisy one.
Training is carried out by considering a large multitemporal SAR image
and its multilook version, in order to approximate a {\em clean} image.
Experimental results, both on synthetic and real SAR data, show the method
to achieve better performance with respect to state-of-the-art techniques.
\end{abstract}

\section{Introduction}
SAR images are affected by a strong multiplicative noise, the speckle,
which may severely impair the performance of automatic operations, like classification and segmentation, aimed at extracting valuable information for the end user.
As more and more images are acquired every day, automatic analysis is mandatory, making of image despeckling a central issue.
A number of approaches have been proposed in the last few years to suppress speckle while preserving relevant image features \cite{Argenti2013}.
Wavelet shrinkage \cite{Bianchi2008}, sparse representations \cite{Foucher2008}, and especially nonlocal filtering \cite{Deledalle2009, Parrilli2012, Cozzolino2014, Deledalle2015},
represent arguably the current state-of-the-art.

\begin{figure}[t!]
	\begin{minipage}[b]{0.48\linewidth} \centerline{\includegraphics[width=\textwidth,trim={120px 210px 10px 0px},clip]{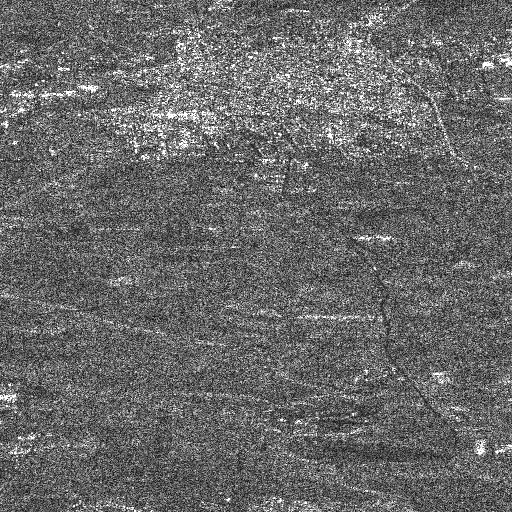}} \end{minipage} \hfill
	\begin{minipage}[b]{0.48\linewidth} \centerline{\includegraphics[width=\textwidth,trim={50px 50px 80px 160px},clip]{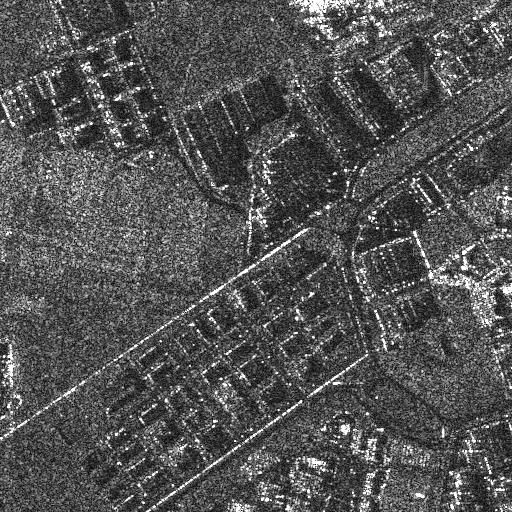}}	\end{minipage}
	\caption{Examples of low-resolution (left) and high-resolution (right) SAR images. Statistical differences appear even by visual inspection.}
	\label{fig:res}
\end{figure}

Most of these approaches rely on detailed statistical models of signal and speckle, either in the original or in a transform domain \cite{Argenti2013}.
However, depending on the sensor, the acquisition modality, the possible use of multilooking, and a number of other factors, including of course the land cover,
statistics may vary significantly from case to case (see Fig.~\ref{fig:res}).
A well-known example concerns high-resolution data such as those acquired by TerraSAR-X, COSMO-SkyMed, and RADARSAT-2 systems.


In this work, we propose to avoid altogether the modeling problem by resorting to the machine learning approach, implemented through a convolutional neural network (CNN).
Given a suitable set of images,
the network is trained to learn an implicit model of the data which allows the effective despeckling of all new data of the same type.
In the last few years, several authors have proposed CNN-based methods for AWGN image denoising
\cite{Jain2009, Burger2012}.
Here, we follow the paradigm proposed in \cite{Zhang2016},
which resorts to residual learning to guarantee a faster convergence in the presence of limited training data.
Adaptation to SAR is obtained by handling multiplicative noise and by using an ad hoc procedure to build a suitable training set.
To the best of our knowledge, this is the first paper investigating CNNs for SAR image despeckling.

In the following sections, we describe the proposed method, we present experimental results on both synthetic and real data, and finally we draw conclusions.

\section{Proposed Method}

\begin{figure*}[!t]
	\centerline{\includegraphics[width=15cm]{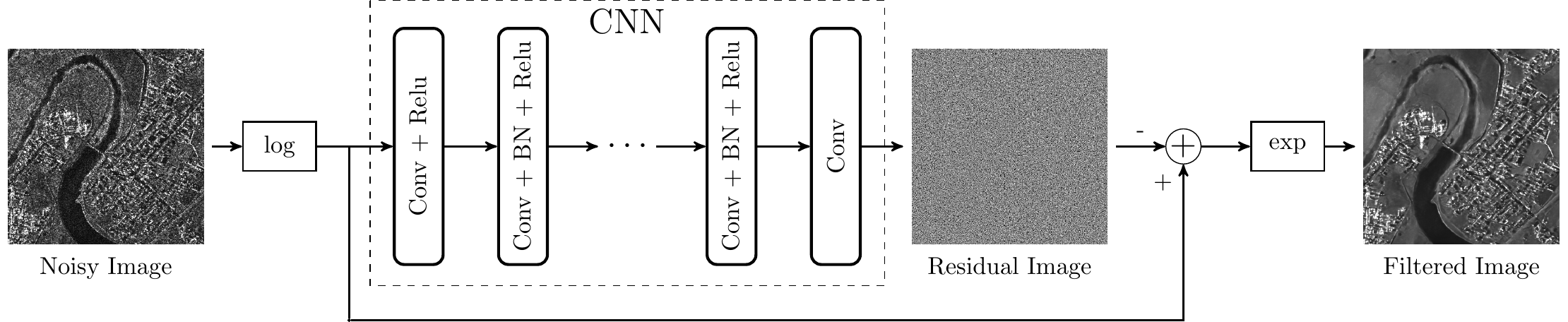}}
	\caption{Proposed CNN architecture for SAR image despeckling.}
	\label{fig:proposed_CNN}
\end{figure*}

The architecture of the proposed CNN, inspired by \cite{Zhang2016}, is shown in Fig.~\ref{fig:proposed_CNN}. The network comprises 17 full convolutional layers, with no pooling.
Each layer extracts 64 feature maps, using filters of size 3$\times$3$\times$64, except the first and last layers which have single-band input and output, respectively. Rather than the clean image, the network recovers the speckle component, which is then subtracted from the noisy image.

A fundamental difference of our approach with respect to \cite{Zhang2016} lies in the criterion to be optimized during training. Indeed, outside of the AWGN realm, the Euclidean distance is not optimal anymore. To deal with multiplicative noise, we use the homomorphic approach with coupled log and exp transforms, in synergy with the similarity measure for speckle noise distribution \cite{Deledalle2012}, leading to the loss function\footnote{$\log$ and $\cosh$ are meant element-wise, whereas $\mathbf{1}^\top z = \sum_{k=1}^K z_k$.}
\begin{equation}
\ell(\mathbf{\Theta}) = \sum_{i=1}^N \mathbf{1}^\top\log\left( \cosh\Big( \mathcal{R}_{\mathbf{\Theta}}(\log \mathbf{y}_i) +c - \log \frac{\mathbf{y}_i}{\mathbf{x}_i} \Big) \right),
\end{equation}
where $\mathcal{R}_{\mathbf{\Theta}}$ is the network output, $\Theta$ denotes the trainable parameters, $(\mathbf{x}_i,\mathbf{y}_i)$ is a pair of clean-noisy training patches in amplitude format, and $c$ is the nonzero mean of log-speckle.

This strategy, called residual learning \cite{He2016}, is key to speed up the training process and helps improving the performance.
In fact, it has been observed experimentally \cite{He2016} that training a CNN may be quite slow when the desired output is very similar to the input.
This is the case of many restoration tasks, such as denoising or super-resolution.
By setting the dual goal of reproducing the noise (hence, removing the clean signal)
training becomes much more effective.
This is extremely important for SAR applications, given the inherent scarcity of training data.
In fact, while for the ``conventional'' case of fully developed speckle one can generate a large dataset by simulation,
this is not possible in other cases, such as high-resolution data, due to the lack of satisfactory models.

\begin{figure*}[!t]
	\centerline{\includegraphics[width=15cm]{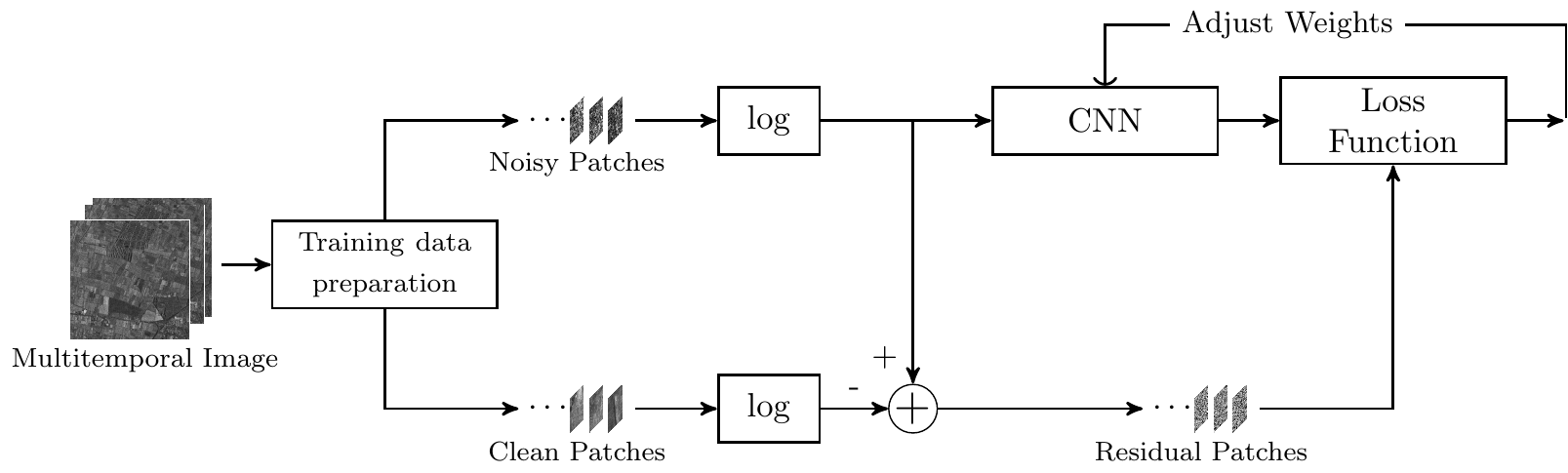}}
	\caption{Training procedure.}
	\label{fig:proposed_training}
\end{figure*}

Therefore, we propose an ad hoc procedure for dataset generation, described graphically in Fig.\ref{fig:proposed_training}.
We assume that a relatively large multitemporal SAR image is available.
The clean image is obtained by averaging the temporal components (multilooking) and keeping only the regions with no significant temporal changes.
Of course, the more temporal instances are available, the more reliable the clean reference is.
Eventually, a number of noisy patches are extracted with their clean version and used to train the network.
The use of residual learning, together with batch normalization and a suitable optimization algorithm \cite{Kingma2015}, allows us to obtain satisfactory training.
The trained network can now deal with any SAR images acquired in the same modality.

\section{Experimental results}\label{sec:experiments}

In SAR image despeckling, the performance assessment is quite challenging,
due to the lack of original noiseless signal.
Therefore, we split the numerical validation in two parts.
First, we present experiments carried out on synthetic SAR images corrupted by simulated speckle,
and make comparison using the usual performance indexes,
such as the peak signal-to-noise ratio (PSNR) and the structural similarity (SSIM).
Afterwards, we experiment with real-world SAR images, focusing on the challenging high-resolution case.

We compare results with three despeckling algorithms, PPB \cite{Deledalle2009}, SAR-BM3D \cite{Parrilli2012}, and NL-SAR \cite{Deledalle2015},
chosen for their competitive performance and the availability of software code.
For all these algorithms parameters are set as suggested in the reference papers.
Turning to the proposed method, called SAR-CNN from now on,
a training set of 2000$\times$128 patches (40$\times$ 40 pixels) is used, with the ADAM gradient-based optimization method \cite{Kingma2015}, minibatches of 128 patches, and the batch normalization strategy of \cite{Ioffe2015}.
Training proceeds for 30 epochs with learning rate 0.001 and, only for synthetic data, further 20 epochs with learning rate 0.0001.
All experiments were carried out in Matlab R2016b with the \emph{MatConvNet} toolbox \cite{Vedaldi2015}, with an Intel Xeon CPU at 2.10GHz and an Nvidia P100 GPU.
Training took about 8 hours.
Interestingly, once training is over, SAR-CNN exhibits the lowest run-time complexity as shown in Tab.\ref{tab:times}.

\begin{table}
	\centering
	\caption{CPU time for despeckling a $512\times512$ clip.}
	\label{tab:times}
	\small
	\begin{tabular}{cccc}
		\toprule
		PPB & NL-SAR & SAR-BM3D & SAR-CNN \\
		\midrule
		49.7 s & 10.1 s & 87.4 s & 4.6 s \\
		\bottomrule
	\end{tabular}
\end{table}

\subsection{Results on simulated SAR images}

\begin{figure*}[t]
	\begin{minipage}[b]{0.19\linewidth}	\centerline{\includegraphics[width=\textwidth,trim={20px 90px 100px 60px},clip]{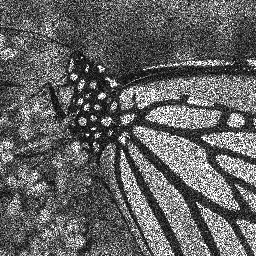}}	    \end{minipage}	\hfill
	\begin{minipage}[b]{0.19\linewidth}	\centerline{\includegraphics[width=\textwidth,trim={20px 90px 100px 60px},clip]{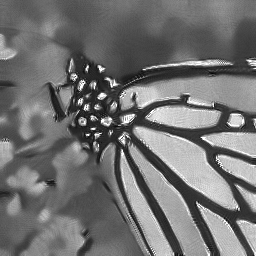}}	    \end{minipage}	\hfill
	\begin{minipage}[b]{0.19\linewidth}	\centerline{\includegraphics[width=\textwidth,trim={20px 90px 100px 60px},clip]{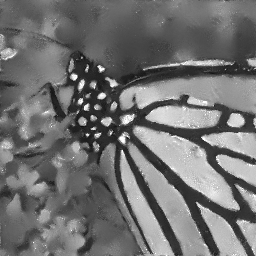}}	    \end{minipage}	\hfill
	\begin{minipage}[b]{0.19\linewidth}	\centerline{\includegraphics[width=\textwidth,trim={20px 90px 100px 60px},clip]{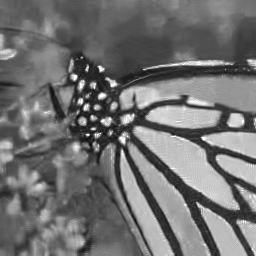}}	\end{minipage}	\hfill
	\begin{minipage}[b]{0.19\linewidth}	\centerline{\includegraphics[width=\textwidth,trim={20px 90px 100px 60px},clip]{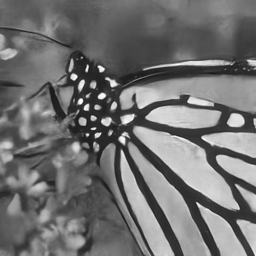}}	\end{minipage}  \hfill
	
	\vspace{1mm} \noindent
	\begin{minipage}[b]{0.19\linewidth}	\centerline{\includegraphics[width=\textwidth,trim={50px 230px 100px 120px},clip]{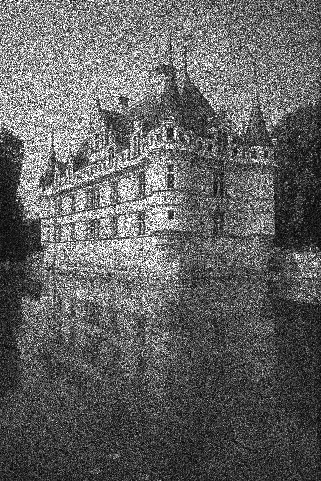}}	    \end{minipage}	\hfill
	\begin{minipage}[b]{0.19\linewidth}	\centerline{\includegraphics[width=\textwidth,trim={50px 230px 100px 120px},clip]{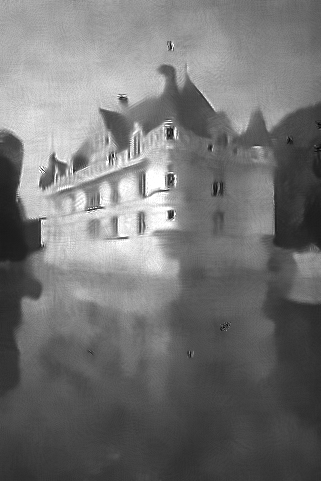}}	    \end{minipage}	\hfill
	\begin{minipage}[b]{0.19\linewidth}	\centerline{\includegraphics[width=\textwidth,trim={50px 230px 100px 120px},clip]{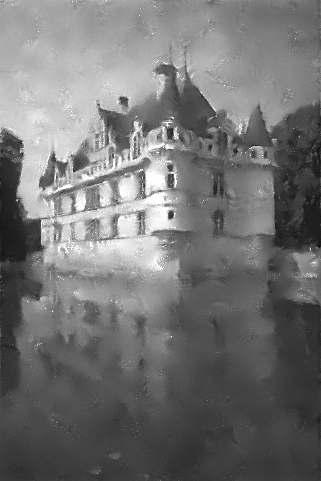}}	\end{minipage}	\hfill
	\begin{minipage}[b]{0.19\linewidth}	\centerline{\includegraphics[width=\textwidth,trim={50px 230px 100px 120px},clip]{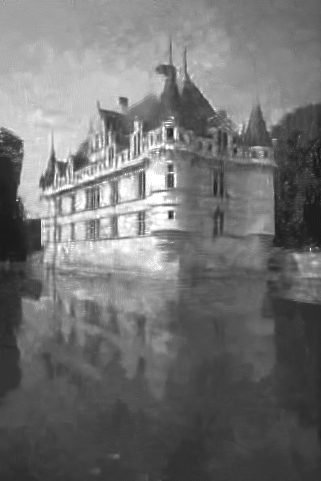}}\end{minipage}	\hfill
	\begin{minipage}[b]{0.19\linewidth}	\centerline{\includegraphics[width=\textwidth,trim={50px 230px 100px 120px},clip]{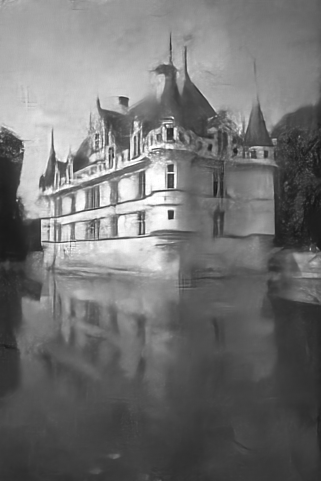}}	\end{minipage}
	\vspace{-1em}
	\caption{Sample results on simulated SAR images. Left to right: original single-look, PPB, NL-SAR, SAR-BM3D, SAR-CNN.
    \label{fig:optical_results}}
\end{figure*}

We generate a number of SAR-like images by injecting single-look speckle in amplitude format on optical images.
Training patches are extracted from 400 different images.
Tab.~\ref{tab:PSNR} reports PSNR results for some out-of-training images often used for testing.
In all but one case SAR-CNN provides the best performance, with an average gain over the reference techniques of about 1 dB, 2 dB, and 2.5 dB, respectively. Similar considerations apply for the SSIM index (Tab.~\ref{tab:SSIM}). Such good results are confirmed by visual inspection, see Fig.~\ref{fig:optical_results}, with an impressive improvement in detail preservation.


\begin{table}[t]
	\centering
	\caption{PSNR over synthetic SAR images.}
	\label{tab:PSNR}
	\footnotesize
	\begin{tabular}{lcccc}
		\toprule
		Image     &  PPB   & NL-SAR & SAR-BM3D & SAR-CNN \\ \midrule
		Cameraman &  23.02 &  24.37 &    24.76 &   26.15 \\
		House     &  25.51 &  25.75 &    27.55 &   28.60 \\
		Peppers   &  23.85 &  23.62 &    24.92 &   26.02 \\
		Starfish  &  21.13 &  21.84 &    22.71 &   23.37 \\
		Butterfly &  22.76 &  23.82 &    24.48 &   26.05 \\
		Airplane  &  21.22 &  21.83 &    22.71 &   23.93 \\
		Parrot    &  21.88 &  24.13 &    24.17 &   25.92 \\
		Lena      &  26.64 &  26.80 &    27.85 &   28.70 \\
		Barbara   &  24.08 &  23.13 &    25.37 &   24.70 \\
		Boat      &  24.22 &  24.55 &    25.43 &   26.05 \\ \midrule
\textsl{Average}  &  23.43 &  23.98 &    24.99 &   \textbf{25.95} \\
		\bottomrule
	\end{tabular}
\end{table}

\begin{table}[t]
	\centering
	\caption{SSIM over synthetic SAR images.}
	\label{tab:SSIM}
	\footnotesize
	\begin{tabular}{lcccc}
		\toprule
		Image     &    PPB & NL-SAR & SAR-BM3D & SAR-CNN \\ \midrule
		Cameraman &  0.661 &  0.716 &    0.750 &   0.792 \\
		House     &  0.651 &  0.686 &    0.751 &   0.791 \\
		Peppers   &  0.680 &  0.716 &    0.747 &   0.793 \\
		Starfish  &  0.563 &  0.609 &    0.664 &   0.702 \\
		Butterfly &  0.714 &  0.752 &    0.792 &   0.841 \\
		Airplane  &  0.533 &  0.620 &    0.672 &   0.724 \\
		Parrot    &  0.685 &  0.732 &    0.771 &   0.805 \\
		Lena      &  0.680 &  0.714 &    0.763 &   0.800 \\
		Barbara   &  0.652 &  0.631 &    0.729 &   0.718 \\
		Boat      &  0.573 &  0.602 &    0.650 &   0.675 \\ \midrule
\textsl{Average}  &  0.639 &  0.677 &    0.729 &   \textbf{0.764} \\
		\bottomrule
	\end{tabular}
\end{table}

\subsection{Results on high-resolution SAR images}

\begin{figure*}[t]
	\begin{minipage}[b]{0.19\linewidth}
		\centerline{\includegraphics[width=\textwidth,trim={50px 50px 80px 160px},clip]{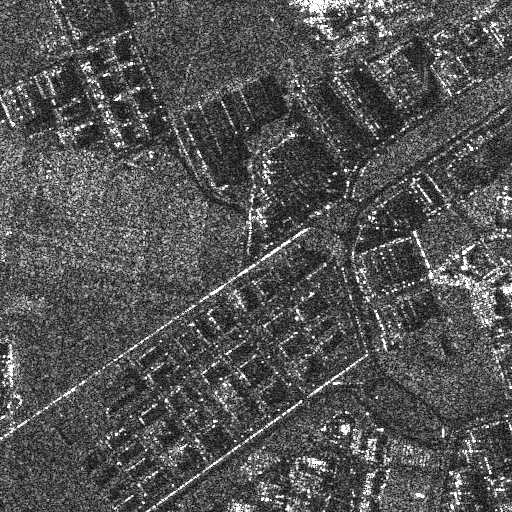}}
	\end{minipage}
	\hfill
	\begin{minipage}[b]{0.19\linewidth}
		\centerline{\includegraphics[width=\textwidth,trim={50px 50px 80px 160px},clip]{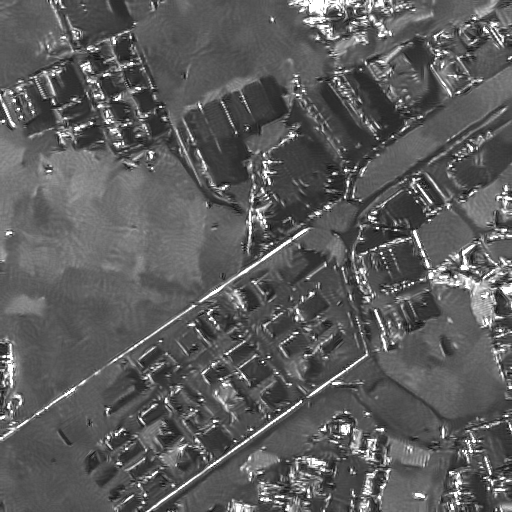}}
	\end{minipage}
	\hfill
	\begin{minipage}[b]{0.19\linewidth}
		\centerline{\includegraphics[width=\textwidth,trim={50px 50px 80px 160px},clip]{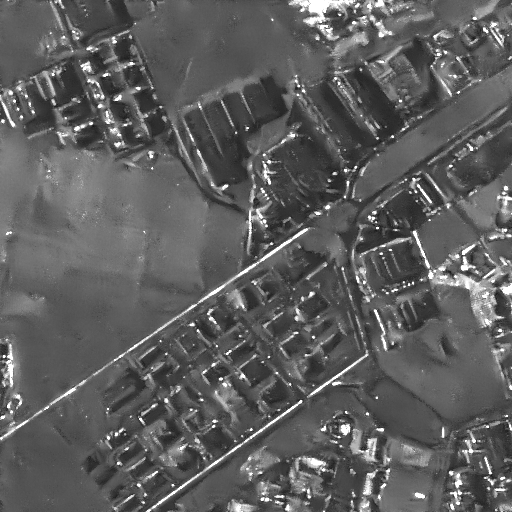}}
	\end{minipage}
	\hfill
	\begin{minipage}[b]{0.19\linewidth}
		\centerline{\includegraphics[width=\textwidth,trim={50px 50px 80px 160px},clip]{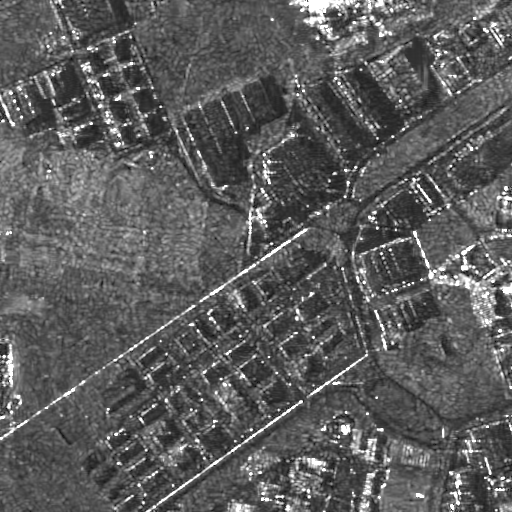}}
	\end{minipage}
	\hfill
	\begin{minipage}[b]{0.19\linewidth}
		\centerline{\includegraphics[width=\textwidth,trim={50px 50px 80px 160px},clip]{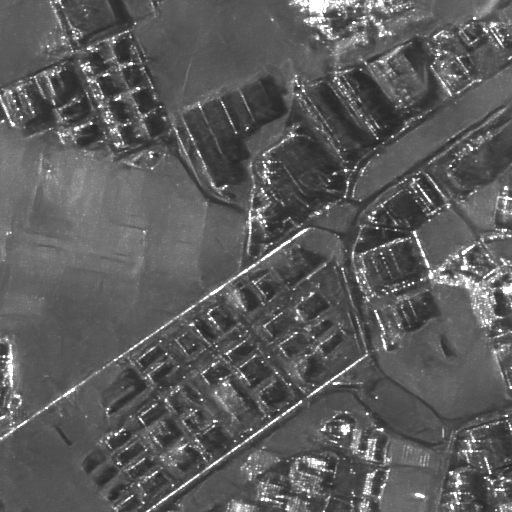}}
	\end{minipage}
	
	\vspace{1mm} \noindent
	\begin{minipage}[b]{0.19\linewidth}
		\centerline{\includegraphics[width=\textwidth,trim={50px 50px 80px 160px},clip]{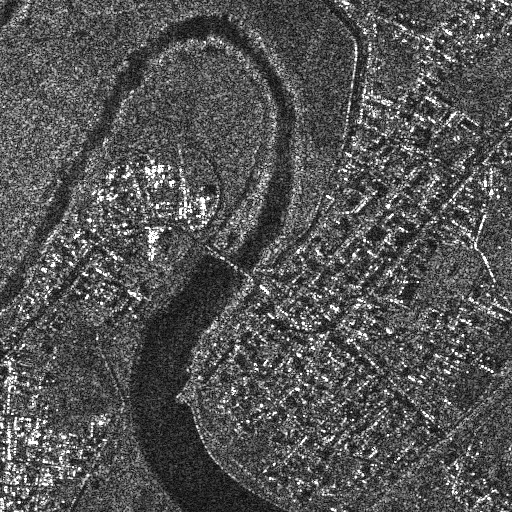}}
	\end{minipage}
	\hfill
	\begin{minipage}[b]{0.19\linewidth}
		\centerline{\includegraphics[width=\textwidth,trim={50px 50px 80px 160px},clip]{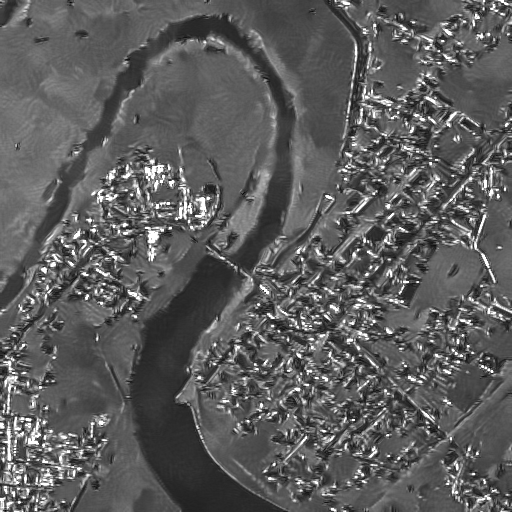}}
	\end{minipage}
	\hfill
	\begin{minipage}[b]{0.19\linewidth}
		\centering
		\centerline{\includegraphics[width=\textwidth,trim={50px 50px 80px 160px},clip]{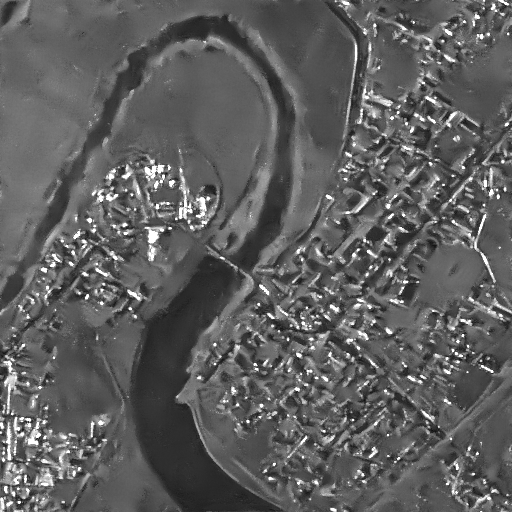}}
	\end{minipage}
	\hfill
	\begin{minipage}[b]{0.19\linewidth}
		\centering
		\centerline{\includegraphics[width=\textwidth,trim={50px 50px 80px 160px},clip]{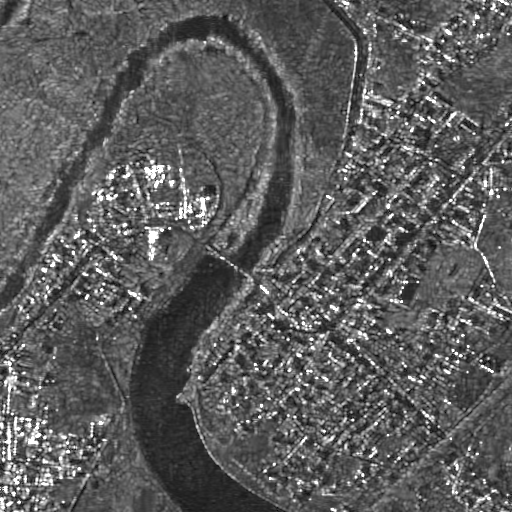}}
	\end{minipage}
	\hfill
	\begin{minipage}[b]{0.19\linewidth}
		\centering
		\centerline{\includegraphics[width=\textwidth,trim={50px 50px 80px 160px},clip]{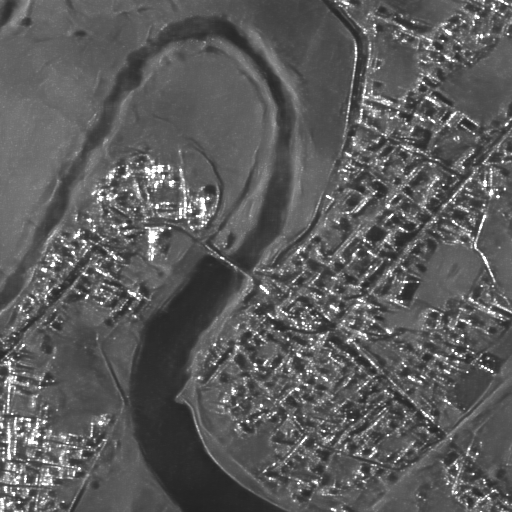}}
	\end{minipage}
	
	\vspace{-1em}
	\caption{Results on a COSMO-SkyMed image. Left to right: original single-look clip, PPB, NL-SAR, SAR-BM3D, SAR-CNN.}
	\label{fig:sar_results}
\end{figure*}

In the second experiment we consider a single-look COSMO-SkyMed image of size 16000$\times$16000 with 25 co-registered temporal components.
Training patches are extracted from one half of the image, whereas numerical evaluation is carried out on several 512$\times$512 clips from the other half.
Fig.~\ref{fig:sar_results} shows results for some of these clips.
Lacking a clean reference, visual inspection is the main tool for quality evaluation.
In our assessment, SAR-CNN looks extremely promising, showing the same speckle suppression ability of NL-SAR, but with a better detail preservation, comparable to that of SAR-BM3D.
These observations are supported also by results of Tab.~\ref{ENL_alfabeta}, reporting two no-reference metrics, the equivalent number of looks (ENL), evaluated on homogeneous blocks,
and the $\alpha\beta$ index \cite{Gomez2016}.
The best scores are achieved by SAR-CNN and NL-SAR, indicating a better speckle suppression and detail preservation.
The improvement w.r.t. competitors is not as striking as in the previous case.
However, note that the network did not see clean patches during training
but only well despeckled ones, with a sure impact on performance.

\begin{table}
\centering
\caption{ENL and $\alpha\beta$ over the clips in Fig.~\ref{fig:sar_results}.}
\label{ENL_alfabeta}
\small
\begin{tabular}{clrrrr}
\toprule
\# clip & Index & PPB & NL-SAR & SAR-BM3D & SAR-CNN \\
\midrule
1 & ENL           & 47.61 & 154.10 &  4.87 & 129.10 \\
2 & ENL           & 25.28 & 52.12 & 4.71 & 56.32 \\
\midrule
1 & $\alpha\beta$ & 0.162 &  0.076 & 0.530 & 0.187 \\
2 & $\alpha\beta$ & 0.171 & 0.065 & 0.511 & 0.182 \\
\bottomrule
\end{tabular}
\end{table}

\section{Conclusion}

In this paper we investigated the use of Convolutional Neural Networks for SAR image despeckling.
A residual learning strategy is applied together with a suitable training phase
that is carried out by using multitemporal SAR data of the same scene
(both the original data and their mutilook version).
Results on synthetic and real SAR data show promising results both considering objective and
visual assessment.

\small
\bibliographystyle{IEEEbib}
\bibliography{biblio_denoising}

\end{document}